\documentclass[
twocolumn,
]{ceurart}

\usepackage{listings}
\usepackage{graphicx}
\usepackage{svg}
\usepackage{subfigure}
\usepackage{caption}
\usepackage{float}
\usepackage{multirow}
\usepackage{subcaption} 
\usepackage{eucal}
\usepackage{lmodern}
\usepackage{hyperref}
\usepackage{cleveref}
\begin{document}

\copyrightyear{2024}
\copyrightclause{Copyright for this paper by its authors. Use permitted under Creative Commons License Attribution 4.0 International (CC BY 4.0)."}

\conference{}

\title{Global Clipper: Enhancing Safety and Reliability of Transformer-based Object Detection Models}

% \tnotetext[1]{You can use this document as the template for preparing your
%   publication. We recommend using the latest version of the ceurart style.}

% \author[1,2]{Dmitry S. Kulyabov}[%
% orcid=0000-0002-0877-7063,
% email=kulyabov-ds@rudn.ru,
% url=https://yamadharma.github.io/,
% ]
\author[1, 3]{Qutub Syed}
\author[1]{Michael Paulitsch}
\author[2]{Karthik Pattabiraman}
\author[1]{Korbinian Hagn}
\author[1]{Fabian Oboril}
\author[1]{Cornelius Buerkle}
\author[1]{Kay-Ulrich Scholl}
\author[3]{Gereon Hinz}
\author[3]{Alois Knoll}
% % \cormark[1]
% % \fnmark[1]
\address[1]{Intel Labs, Munich, Germany}
\address[2]{University of British Columbia, Vancouver, Canada}
\address[3]{Technical University of Munich, Munich, Germany}

% \author[3]{Ilaria Tiddi}[%
% orcid=0000-0001-7116-9338,
% email=i.tiddi@vu.nl,
% url=https://kmitd.github.io/ilaria/,
% \address[3]{Vrije Universiteit Amsterdam, De Boelelaan 1105, 1081 HV Amsterdam, The Netherlands}

% \author[4]{Manfred Jeusfeld}[%
% orcid=0000-0002-9421-8566,
% email=Manfred.Jeusfeld@acm.org,
% url=http://conceptbase.sourceforge.net/mjf/,
% \address[4]{University of Skövde, Högskolevägen 1, 541 28 Skövde, Sweden}

% \cortext[1]{Corresponding author.}
% \fntext[1]{These authors contributed equally.}

\begin{abstract}
As transformer-based object detection models progress, their impact in critical sectors like autonomous vehicles and aviation is expected to grow. Soft errors causing bit flips during inference have significantly impacted DNN performance, altering predictions. Traditional range restriction solutions for CNNs fall short for transformers. This study introduces the Global Clipper and Global Hybrid Clipper, effective mitigation strategies specifically designed for transformer-based models. It significantly enhances their resilience to soft errors and reduces faulty inferences to ~ 0\%. We also detail extensive testing across over 64 scenarios involving two transformer models (DINO-DETR and Lite-DETR) and two CNN models (YOLOv3 and SSD) using three datasets, totalling approximately 3.3 million inferences, to assess model robustness comprehensively. Moreover, the paper explores unique aspects of attention blocks in transformers and their operational differences from CNNs.
\end{abstract}

% \begin{keywords}
%   LaTeX class \sep
%   paper template \sep
%   paper formatting \sep
%   CEUR-WS
% \end{keywords}

\maketitle

\section{Motivation}
\label{sec:intro}

The adoption of Deep Neural Networks (DNNs) has significantly impacted various sectors, including autonomous vehicles \cite{wang2020safety}, aviation, healthcare \cite{habli2020artificial}, and space exploration \cite{oche2021applications}, where high safety and reliability are crucial. This has spurred the growth of computer vision research communities focused on safe AI, tackling areas such as out-of-distribution detection \cite{gawlikowski2023survey}, adversarial robustness and model interoperability \cite{xu2019explainable}. A DNN-based computer vision model processes images to classify objects and predict their bounding boxes.

\begin{figure}[b]
  \centering
  \includegraphics[width=\linewidth]{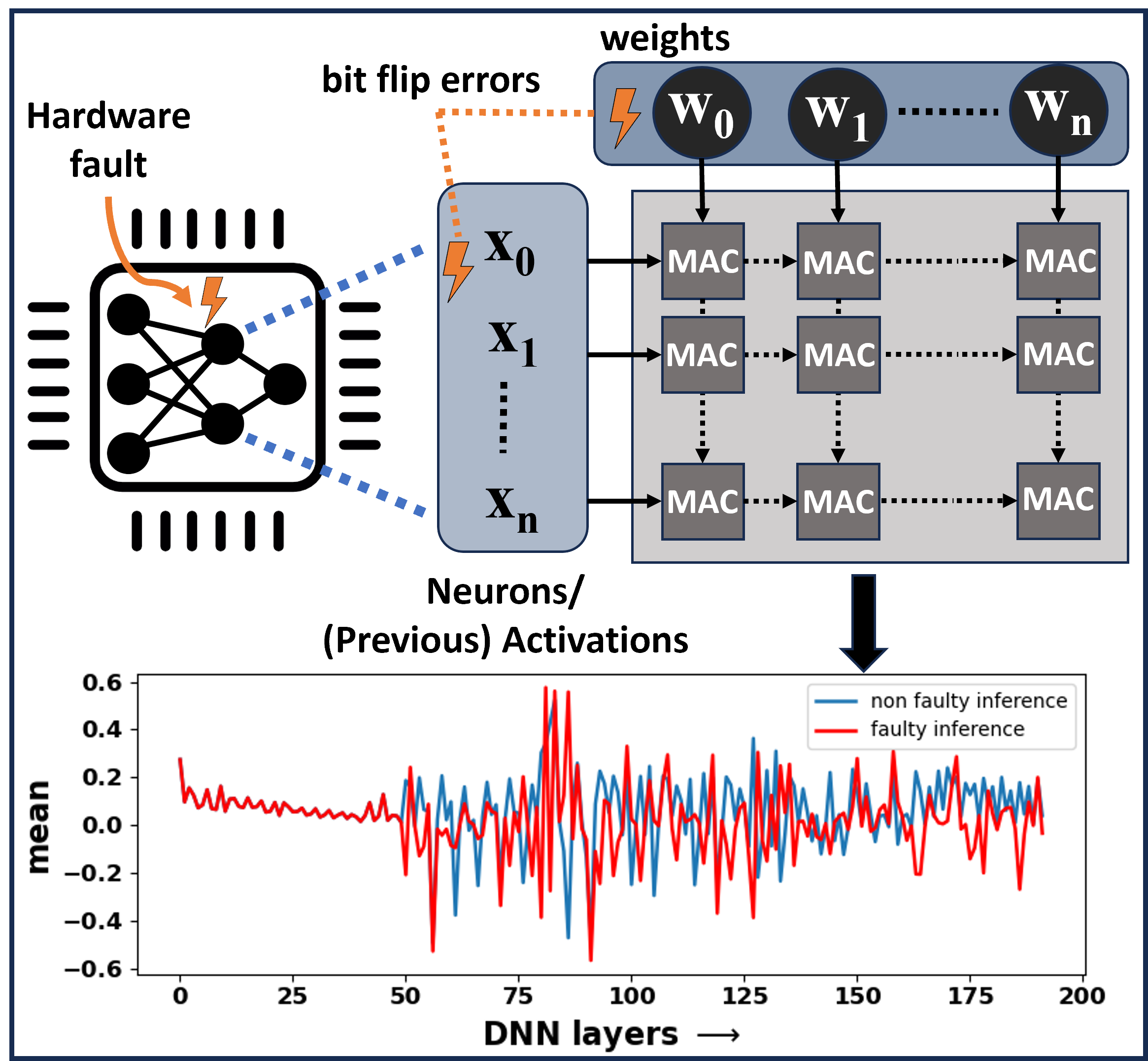}
  \caption{Abstract architecture of a DNN accelerator. The upper figure illustrates potential soft errors resulting in bit flips within neurons or weights at specific layers of the DNN model. The lower figure displays the mean values of layers in non-faulty inferences compared to faulty inference values when a bit-flip error is injected at the 50th layer of a transformer model DINO-DETR.}
  % \vspace{-0.45cm} 
  \label{fig:abstract_dnn_accelerator}
\end{figure}

Errors during inference can lead to faulty bounding boxes, significantly altering system behaviour and underscoring the critical need for safer hardware for model execution. DNN accelerators execute models at a high level by constructing a computational graph that uses General matrix-to-matrix multiplication (GEMM) \cite{gao2023systematic} for extensive layer input and weight multiplications. Key components in this process are the Multiply-accumulate (MAC) units within the lower accelerator levels shown in \cref{fig:abstract_dnn_accelerator} \cite{chen2020survey}. %, lack ECC protection, unlike DRAMs, making them particularly vulnerable. 
MAC units in DNN accelerators lack ECC protection, making them particularly vulnerable to soft errors—a major reliability concern. Such errors, often caused by radiation, chip ageing, manufacturing variations, or thermal issues \cite{mukherjee2005soft, 5493404, ibrahim2020soft}, can alter intermediary computational values, leading to incorrect inferences. Research shows that the soft error rate will increase with higher transistor density, reduced feature sizes, and more cores \cite{upasani2016soft, borkar2005designing, shivakumar2002modeling}. For example, a 100-core system  of 16nm node may fail every 1.5 hours due to soft errors \cite{upasani2016soft}, significantly affecting predictions as these propagate through layers, as shown in \cref{fig:abstract_dnn_accelerator}. Although soft errors do not cause permanent damage, they can result in substantial reliability degradation. 
% Furthermore, the work on soft error mitigation \cite{upasani2016soft} shows that the soft error rate of current and future processors is expected to increase exponentially because of the exponential growth rate of on-chip transistors, the shrinking feature size, and increasing core count \cite{borkar2005designing, shivakumar2002modeling}. They have computed the soft error rate which says a system of 100 cores/chip will face a failure every 1.5 hours.

% Dependence on safe and reliable technology necessitates comprehensive testing across multiple levels and paradigms, with subsequent consolidation of measured metrics into a unified category \cite{billinton1992reliability}. 

% The DNNs (Deep Neural Networks) have been demonstrated to be affected by these soft errors, as evidenced by the literature \cite{geissler2021towards, geissler2023low, asgari2023structural, qutub2022hardware}. These studies have shown instances where single-bit flip errors have led to faulty inferences. Moreover, besides the impact of soft errors on DNN predictions, there is additional research on bit-flip attacks \cite{rakin2019bit, li2020defending}, indicating the susceptibility of DNNs to perturbations. 
This paper proposes a technique for mitigating soft errors in object detection models at the application level. We simulated soft errors as bit flips using PytorchALFI \cite{grafe2023large}, an open-source tool that integrates large-scale fault injection capabilities with PyTorch.

Range restriction solutions effectively mitigate soft errors in CNN-based DNN models by applying pre-calculated bounds at every activation layer, computed using 20\% of validation images to determine the minimum and maximum restrictions \cite{chen2021low}. However, current protective measures fall short against soft errors in transformer-based vision models due to the complexity of their architectures. Our analysis shows that existing solutions are inadequate, necessitating significant enhancements in error mitigation strategies for these advanced systems. Without such improvements, the robustness of transformer-based models is compromised, highlighting the urgent need for more sophisticated and tailored protection mechanisms. Transformer models \cite{zhang2022dino, li2023lite}, characterized by their self-attention and large linear layers, are particularly vulnerable, as bit flip errors can cascade and significantly alter predictions.

\begin{figure}
  \centering
  \includegraphics[width=\linewidth]{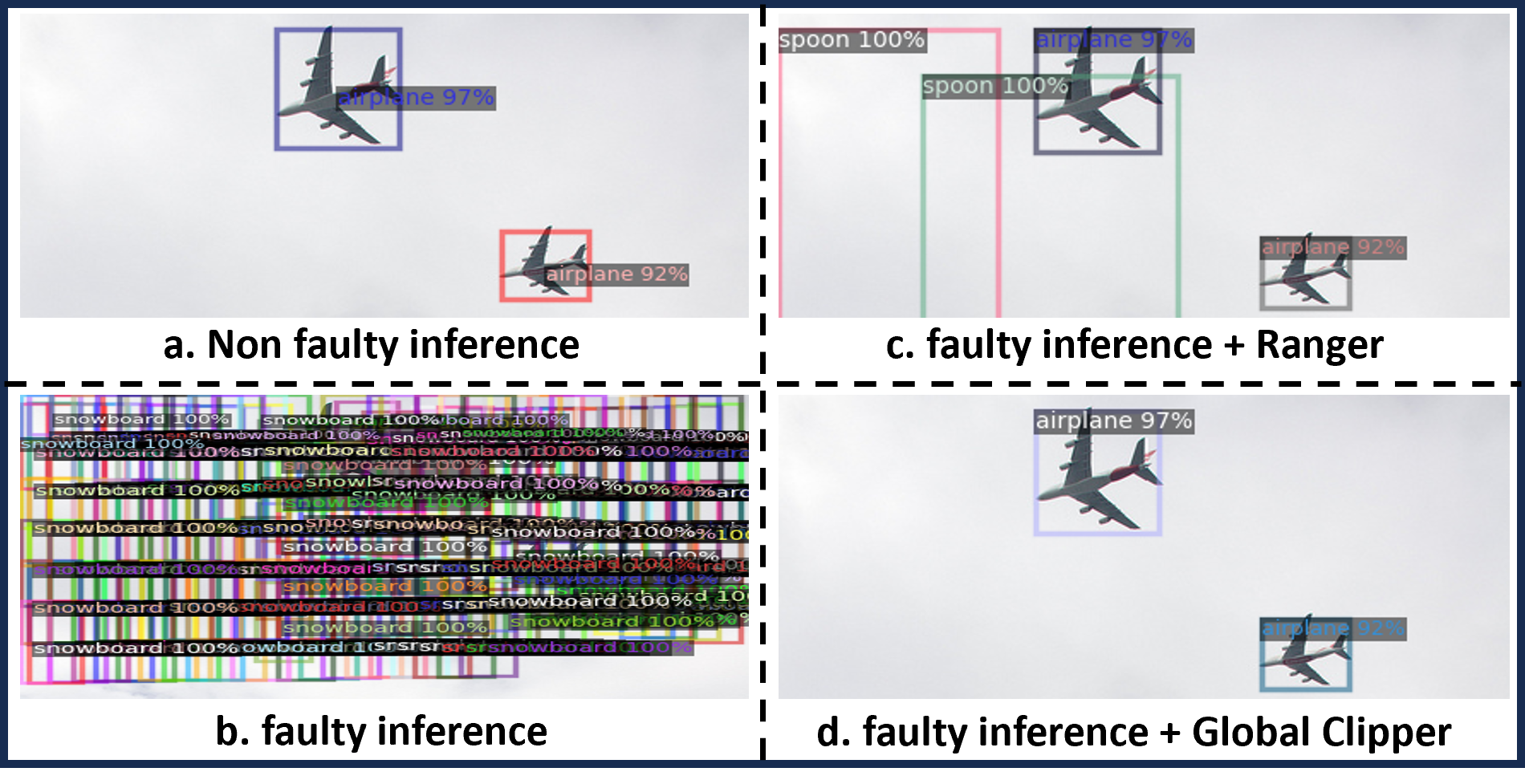}
  \caption{Visual example of faulty inferences on CoCo trained DINO-DETR model due to bit-flips caused by the soft errors.}
  \label{fig:faulty_inference_example}
  % \vspace{-0.45cm} 
\end{figure}

For example, injecting a single-bit flip error into the 50th self-attention layer of the CoCo-trained DINO-DETR model, with 48M parameters \cite{zhang2022dino}, results in faulty inference such as ghost objects as shown in \cref{fig:faulty_inference_example}b. These errors, which either create random high-confidence detections or erase them, can disrupt systems dependent on these models for tasks like object tracking, as shown in \cite{qutub2022hardware}. This underscores the significant impact of minor errors in complex networks. However, applying existing range restrictions cannot mitigate all ghost objects, as illustrated in \cref{fig:faulty_inference_example}c.

We propose the Global Clipper and Global Hybrid Clipper range restriction layers as a straightforward yet vital enhancement to mitigate the impacts of soft errors in complex transformer-based models. These layers are  implemented within the activation and linear layers of self-attention blocks, crucial points vulnerable to errors that can drastically affect network performance. This strategy involves a nuanced balance: preserving the network’s ability to process diverse data inputs while ensuring robustness against errors that could lead to significant inaccuracies in outputs. By adding these range restriction layers, Global Clipper effectively safeguards the network’s functionality without compromising its learning capabilities, ensuring sustained high performance even under challenging conditions. In the example, when Global Clipper is added, all the false ghost objects created by fault injection are removed as shown in \cref{fig:faulty_inference_example}d. 

In summary, the main contributions of this paper are as follows:
\begin{enumerate}
    \item \label{contrib:first} We introduce the Global Clipper and Global Hybrid Clipper fault mitigation techniques for transformer-based object detection models (\cref{sec:method}). 
    \item \label{contrib:second} We present a comprehensive study with fault injection experiments across CNN and transformer models using three datasets, totalling 3.3 million inferences, to analyze vulnerabilities in these vision systems. We show that the proposed techniques are effective in reducing error rates from ~6\% to nearly 0\% (\cref{subsec:global_clipper_results}).
    \item  \label{contrib:third} We explore the unique characteristics of attention blocks in transformers versus CNNs, providing insights into model vulnerabilities essential for enhancing safety throughout the life-cycle of deployed transformer models (\cref{subsec:ablation_study}).

    % \item Our extensive experiments demonstrate that transformer-based object detection models exhibit greater inherent resiliency against soft errors than CNN-based object detection models.
\end{enumerate}

The following sections of the paper will first explore established methods for addressing soft errors in transformer-based DNN models. Then, the paper will detail the fault injection models under consideration ( as discussed in above contributions - \cref{contrib:second}). Next, our proposed solution, Global Clipper (\cref{contrib:first}), will be introduced, emphasizing its effectiveness in mitigating soft errors. Subsequently, a thorough ablation study on implementing Global Clipper will be conducted(\cref{contrib:second}). Finally, an examination and comparison of the vulnerability characteristics of Transformers and CNN across diverse datasets will be provided (\cref{contrib:third}).

% Ranger restriction solutions for CNN-based models focus on monitoring and restricting activation layers. This approach is effective because convolution layers operate on local attention, concentrating on specific parts of the image or feature maps. Thus, even if a bit flip occurs at a higher order, constraining values at activation layers ensure that further layer computations confine any deviations to a local spatial patch of feature maps.
% In contrast, vision models \cite{dosovitskiy2020image} based on self-attention layers consist of large linear layers. With their global attention mechanism, bit-flip errors in these layers can propagate throughout the network, affecting all vectors in multi-head attention layers and altering predictions. 
% \subsection{Safety issues of computer vision models}
% In this version of the BFA, the attacker randomly selects a bit in any layer of a DNN model. It requires no knowledge of the model. This attack needs flipping of at least 10\% of the weights to be able to break down the model as reported in [12] (J. Li, M. Mao and C. Chakrabarti, "Improving the reliability of rearm-based on implementation through novel weight distribution", 2019 IEEE International Workshop on Signal Processing Systems (SiPS), pp. 189-194).

% a. Adversarial attacks, 
% b. safety of DNN
% c. 
% \subsection{Hardware-related issues causing faulty inferences}

\section{Related Work}
\label{sec:related_work}
The reliability of safety-critical DNN models is assessed through various metrics at the application and hardware levels, enhancing their safety and reliability \cite{neale2016neutron, li2017understanding, chen2021low, qutub2022hardware}. Research has shown that DNNs are prone to soft errors, with instances of single-bit flips leading to faulty inferences \cite{geissler2021towards, geissler2023low, asgari2023structural, qutub2022hardware}.
Traditionally, protection from soft errors in hardware has primarily involved error detection or correction codes (EDC or ECC) \cite{hamming1950error} for memory and using residuals for computing. These mechanisms are commonly implemented in high-end server-grade CPUs but less so in GPUs due to cost considerations or typical relaxed application requirements. Other techniques based on redundancy, like DMR (dual modular redundancy) and TMR (triple modular redundancy), are also used. Despite these measures, accelerators running DNNs may lack inherent protection. 
Furthermore, modular redundancy techniques can be implemented through ensembles, as discussed by \cite{shlezinger2021collaborative}, which may be used to detect and mitigate faults. However, these methods come with substantial computational overhead. To address this issue, budding ensemble solutions \cite{qutub2023bea, Qutub_2024_CVPR} could be explored to reduce the computing overhead. Despite their potential, these solutions have not yet been demonstrated for mitigating or detecting soft errors.

Typical solutions for matrix operations in software, like algorithm-based fault tolerance (ABFT), have been adapted for DNNs but are limited by the overhead of checking large matrix multiplications typical in DNN applications \cite{chen2013online, zhao2020ft}. Researchers have also developed DNN-specific solutions at the application level based on range restriction solutions at the software level, particularly for CNN models, and explored using activation patterns to detect soft errors \cite{hoang2020ft, chen2021low, zhao2020ft}. A small machine-learning module, reduced in dimension, analyses these patterns to identify and reject erroneous inferences \cite{schorn2020facer, schorn2018efficient, zhao2022uniform, mahmoud2020hardnn}. However, these methods face challenges related to scalability and complexity.

The vulnerability of DNNs to soft errors, including a significant number of CNN and few transformer-based models, is well-documented \cite{ibrahim2020soft, li2017understanding, qutub2022hardware, xue2023soft, roquet2024cross, geissler2023low}. However, previous studies have not extensively explored transformer models in object detection or conducted detailed, large-scale fault injection studies \cite{agarwal2023resilience, roquet2024cross}. This study aims to fill this gap by examining the resilience of transformer architectures and exploring effective mitigation strategies against soft errors in these architectures.
% roquet2024cross

\section{Fault Models}
\label{sec:fault models}
\begin{figure*}
\centering
\includegraphics[width=\linewidth]{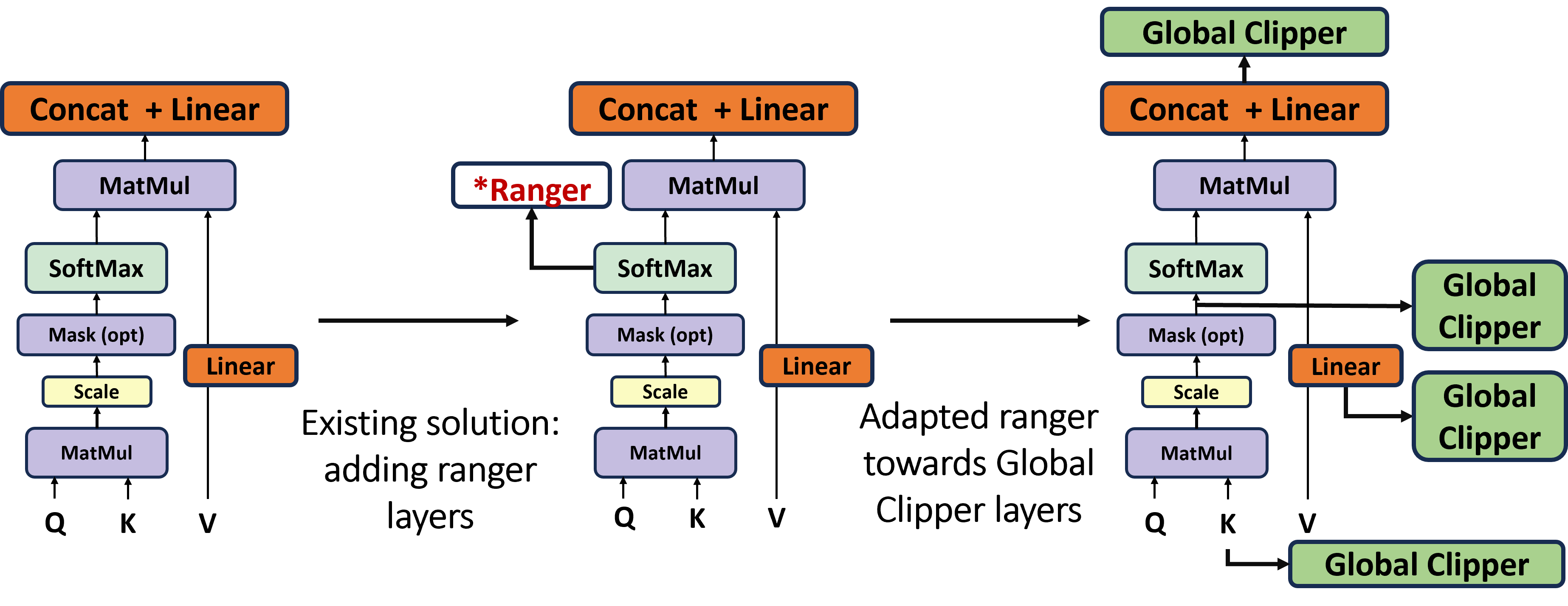}
\caption{Integrating Global Clipper layers into transformer-based object detection models' self-attention blocks. *Ranger layers are recommended to be added to activation functions, usually at ReLU layers, not SoftMax.}
% \vspace{-0.45cm}
\label{fig:intro_global_clipper}
\end{figure*}

We consider soft errors in AI hardware accelerators, focusing on their impact on system reliability. These errors, typically manifesting as single or multiple-bit flips, can compromise data integrity by altering the model's weights and neurons, potentially skewing computations and decisions. Such disruptions in deep neural network operations are illustrated in \cref{fig:abstract_dnn_accelerator}, highlighting the need for robustness strategies.

Parity or Error-Correcting Code (ECC) protects memory against soft errors, particularly crucial for essential memory blocks due to the significant overhead \cite{Lotfi2019}. While ECC, especially SECDED (Single Error Correct, Double Error Detect) code, can detect and correct single-bit errors, it is limited to detecting two-bit errors without correction. This %reveals ECC’s restrictions in handling multi-bit flips and 
underscores the necessity for %its selective application, 
other techniques beyond ECC, 
where minimizing multi-bit errors is essential.

Our experimental setup injects faults as single or 10-bit flip errors during inference, with isolated injections in either the neurons or weights of the model, but not both simultaneously. This method ensures targeted and straightforward fault analysis, with each inference undergoing a single fault alteration. All models in our study employ 32-bit data types. %, focusing on effectively managing and mitigating soft errors.

\section{Global Clipper}
\label{sec:method}
Range restriction solutions \cite{chen2021low, hoang2020ft} effectively address bit flips caused by soft errors in CNN-based models by focusing on activation layers where convolutional layers attend to local image areas. This containment of deviations within localized feature map areas helps prevent extensive errors. However, these methods are less effective for transformer models, which employ global attention mechanisms across extensive linear layers \cite{dosovitskiy2020image}. In transformers, a bit flip can propagate errors throughout the multi-head attention layers, significantly altering vector representations and impacting predictions. This necessitates different mitigation strategies tailored to the global processing nature of transformers.

We introduce a crucial enhancement to existing range restriction layers, as illustrated in \cref{fig:intro_global_clipper}, extending value monitoring and truncation from activation layers to linear layers within self-attention blocks. This strategy bolsters transformer architectures against soft-error-induced bit flips. The Global Clipper truncates out-of-range values to a predefined interval before deployment, operating at any activation or linear layer $\mathcal{l}$ with bounds \(B_{\text{lower}}, B_{\text{upper}}\), as detailed in \cref{eq:global_clipper_eq}. Similarly, the Global Ranger restricts values within specified bounds, ensuring all layer outputs adhere to expected ranges.

\begin{figure}[b]
  \centering
  \includegraphics[width=\linewidth]{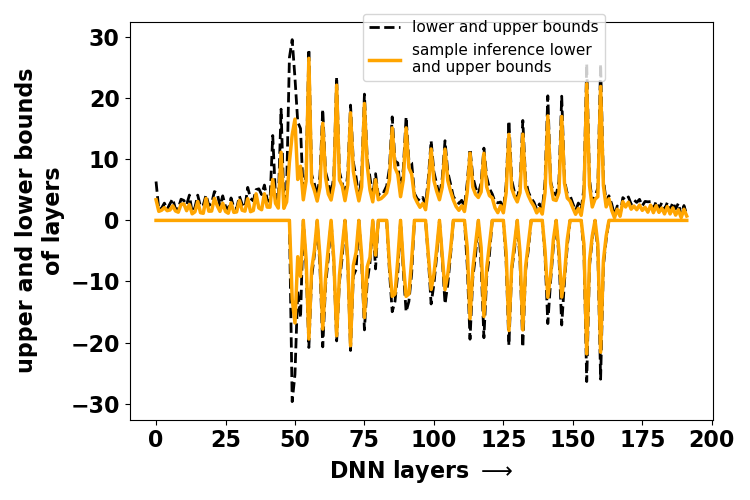}
  \caption{Lower and upper bounds for range restrictions, encompassing activation layers and linear layers within the self-attention blocks of the DINO-DETR model, are defined by the Global Clipper technique.}
  \label{fig:global_clipper_range_bounds}
% \vspace{-0.45cm}
\end{figure}

These layers can be seamlessly fused at the application level, ensuring minimal overhead. Determining upper and lower bounds follows the approach outlined in previous range restriction solutions like Ranger \cite{chen2021low}. Specifically, these bounds are computed using 20\% of the training dataset, encompassing all activation and linear layers within transformer-based models as shown in the \cref{fig:global_clipper_range_bounds}. 
 % \( A_{\text{sub}} \)

\begin{equation}
% \vspace{-25pt}
% \begin{split}
            L_{global\_clipper}(x) =
\begin{cases}
    0 & \text{if}~x <  B_{lower}~or~x >  B_{upper}\\
    0 & \text{if}~\text{Inf}\ \lor\  \text{NaN}\\
    x &  otherwise
\end{cases}
% \end{split}
% \vspace{-25pt}
\label{eq:global_clipper_eq}
\end{equation}
% \vspace{0.5pt}

\begin{figure}
\centering     %%% not \center
\subfigure[Fault injected in ReLU layer]{\label{fig:track_mean_var_relu}\includegraphics[width=\linewidth]{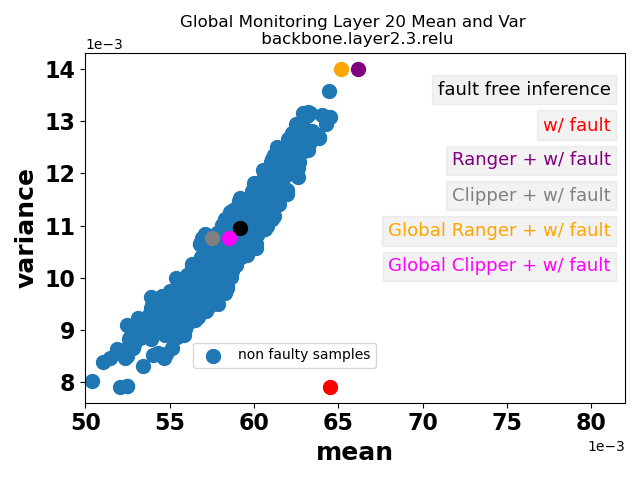}}
\subfigure[Fault injected in linear layer]{\label{fig:track_mean_var_linear}\includegraphics[width=\linewidth]{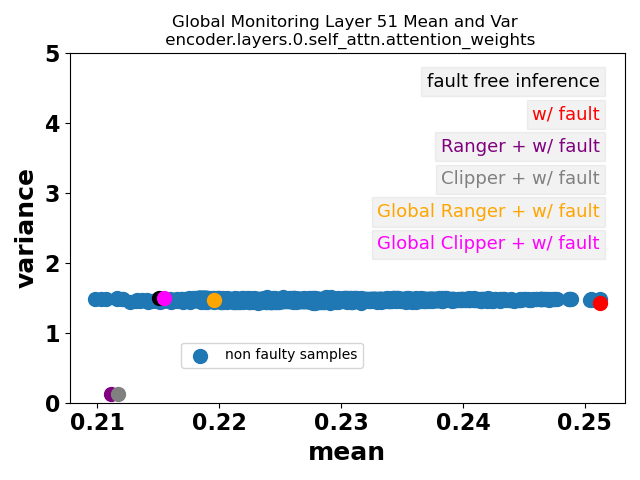}}
\caption{Tracking the mean and variance of layers within the DINO-DETR model, this illustration focuses on the ReLU activation layer and linear layer within the self-attention block.}
% \vspace{-0.45cm}
\label{fig:track_mean_var}
\end{figure}

We demonstrate the effectiveness of the Global Clipper solution using two experiments that involve injecting faults into the convolution and linear layers of existing models alongside our proposed method. Using the sample images from validation datasets, we introduce bit flip errors at a random MSB position in the second convolution layer of the ResNet50 backbone within the DINO-DETR model. We then monitor the mean and variance at the 20th ReLU activation layer and the 51st linear layer in the self-attention block immediately following the ResNet50 encoder. This process, illustrated in \cref{fig:track_mean_var}, allows us to evaluate the performance of the Global Clipper and compare it with other mitigation strategies.

In the first experiment, we inject faults into the ReLU activation layer and apply various mitigation solutions, as shown in \cref{fig:track_mean_var_relu}. The Ranger method \cite{chen2021low} confines data points within the extrapolated population cluster space. In contrast, the Clipper \cite{hoang2020ft}, along with the proposed Global Clipper, more tightly constrains the data to the region of non-faulty inferences. Although all techniques maintain acceptable tolerances, Ranger and Clipper tend to shift data points further from the original non-faulty positions at linear layers.

In our second experiment, we inject faults into the linear layer and activate various mitigation solutions. However, neither Ranger nor Clipper can confine the values to their original positions due to the sensitivity of self-attention layers to faults. Both Ranger and Clipper typically apply restrictions only at the activation layer; however, self-attention includes a SoftMax layer that cannot be similarly restricted without impairing the functionality and accuracy of the block. Hence, introducing the Global Clipper, shown in \cref{fig:intro_global_clipper}, is essential for protecting attention blocks from faults. With faults introduced into the 51st linear layer, the Global Clipper successfully maintains data points closer to their original positions, as demonstrated in \cref{fig:track_mean_var_linear}.
While the definition of Global Clipper resembles Ranger's, refining the bounds is crucial as it ensures that feature values remain within the sampled space in specific layers. In both experiments mentioned above, fault injection demonstrated using sample data doesn't necessarily result in faulty predictions. However, fault locations are sampled to visualize all data points within the area, providing a clear explanation of Global Clipper. Additionally, more faults affect the layer values, causing them to shift into log space compared to the fault-free visualizations. There are a few cases in which Global Clipper may not function out of the box on certain models, requiring slight modifications. These exceptions and adaptations are further elaborated upon in \cref{subsec:global_clipper_results}.

\section{Experiments}
\label{sec:experiments}

\subsection{Experimental Setup}
\label{subsec:experimental setup}
% We estimate the vulnerabilities of the models by introducing bit flips at the application layer, focusing on two CNN-based Object detection models \cite{redmon2018yolov3, liu2016ssd} across 3 datasets: CoCo \cite{lin2014microsoft}, KITTI \cite{geiger2012we} and BDD100K dataset \cite{yu2020bdd100k}. Additionally, we evaluate two transformer-based object detection models DINO-DETR \cite{zhang2022dino} and Lite-DETR \cite{li2023lite} trained on CoCo, KITTI and BDD100K datasets.
As introduced in \cref{sec:intro}, soft errors, characterized by transient bit flip errors at the application level, impact individual inferences and last only until the next data fetch from memory. Bit flips at the sign and most significant bits of the mantissa minimally affect value ranges; however, flips at the sign and exponent bits of IEEE 754 floating-point arithmetic can alter predictions, which does not significantly change when testing other formats like BFloat16 as seen in \cite{geissler2021towards, qutub2022hardware}.

Our study assesses vulnerabilities in two transformer-based models, DINO-DETR \cite{zhang2022dino} and Lite-DETR \cite{li2023lite}, and two CNN models, YOLOv3 \cite{redmon2018yolov3} and SSD \cite{liu2016ssd}, across the CoCo \cite{lin2014microsoft}, KITTI \cite{geiger2012we}, and BDD100K datasets \cite{yu2020bdd100k}. We conducted over 64 experiments, totalling about 3.3 million inferences. Each model undergoes experiments with random and targeted fault injections across all linear layers of self-attention blocks for transformers and convolution layers for CNN models.

Each data point extracted from these experiments includes 50,000 inferences from 1,000 image samples with random faults and 10,000 inferences per targeted fault experiment at each layer. Each experiment set repeats with baseline and proposed mitigation techniques, like Clipper, Global Clipper or Global Hybrid Clipper, allowing a thorough analysis of model vulnerabilities and the effectiveness of mitigation strategies.

\subsection{Evaluation Metrics}
\label{subsec:evaluation_metrics}

Accuracy metrics like AP50 or mAP \cite{lin2014microsoft} and their variants \cite{dos2019impact} are standard for evaluating fault injections in object detection models. Occasionally, these faults create ghost objects with lower confidence scores, not affecting the overall AP50 due to their exclusion in the PR curve area under curve (AUC) calculations \cite{qutub2022hardware}. To address this issue, the $IVMOD$ metric \cite{qutub2022hardware}, which is insensitive to PR curve averaging, is employed. Faults that do not alter the model's outcome are considered benign. In contrast, significant faults are categorized into SDC (silent data corruption) and DUE (detectable and unrecoverable error) as recognized by the safety and reliability community \cite{mukherjee2005soft}.

% \begin{figure}[h]
% \centering
% \includegraphics[width=0.9\linewidth]{images/archs/global_clipper.png}
% \caption{Introduction of Global Clipper layers into self-attention blocks of transformer-based object detection models. Ranger layers are recommended to be added to activation functions, usually at ReLU layers, not SoftMax.}
% % \vspace{-0.45cm} 
% \label{fig:intro_global_clipper}
% \end{figure}

By defining SDC and DUE \cite{mukherjee2005soft} as critical faults, we enhance our ability to assess vulnerabilities effectively. This study introduces these conditions into the $IVMOD_{fd}$ metric for faulty detections (see \cref{eq:SDC_orig}, \cref{eq:DUE_orig}, and \cref{eq:SDC_DUE_compressed}). Unlike the Global Clipper, the Global Ranger restricts values without truncating them. Vulnerability is evaluated by monitoring AP50 accuracy and $IVMOD_{fd}$. For example, if 30 out of 100 sampled images show detection discrepancies or encounter $NaN$ or $inf$ errors due to bit flips, the $IVMOD_{fd}$ would be 30\%. Additionally, $IVMOD_{fd}$ can be used to estimate DNN accelerator vulnerability in terms of FIT rates \cite{mukherjee2005soft} and other risk factors throughout the hardware’s lifecycle \cite{neale2016neutron, li2017understanding}. However, these aspects are beyond this paper’s scope. Hereafter, $IVMOD_{fd}$ will be interchangeably called faulty detections. Moreover, in this study, $IVMOD_{fd}$ solely considers the 9 higher-order bits, including the sign and exponent bits, as described in \cref{sec:fault models}.

\begin{equation}
% \vspace{-25pt}
\begin{split}
            % IVMOD_{fd} = IVMOD_{SDC} \lor\ IVMOD_{DUE}, \\
        IVMOD_{SDC} = \frac{1}{N} \sum^{N}_{i=1} [ (FP_{\text{orig}})_i \neq (FP_{\text{corr}})_i \lor\ \\ (FN_{\text{orig}})_i  \neq (FN_{\text{corr}})_i\big] \\
        % DUE &= \frac{1}{N} \sum^{N}_{i=1} \left[ \text{Inf}_i\ \lor\ \text{NaN}_i \right].
\end{split}
% \vspace{-25pt}
\label{eq:SDC_orig}
\end{equation}

\begin{equation}
% \vspace{-25pt}
\begin{split}
            % IVMOD_{fd} = IVMOD_{SDC} \lor\ IVMOD_{DUE}, \\
        IVMOD_{DUE} = \frac{1}{N} \sum^{N}_{i=1} [\text{Inf}\ \lor\  \text{NaN}] \\
        % DUE &= \frac{1}{N} \sum^{N}_{i=1} \left[ \text{Inf}_i\ \lor\ \text{NaN}_i \right].
\end{split}
% \vspace{-25pt}
\label{eq:DUE_orig}
\end{equation}

\begin{equation}
% \vspace{-5pt}
\begin{split}
            IVMOD_{fd} = IVMOD_{SDC} \lor\ IVMOD_{DUE}, \\
\end{split}
% \vspace{-25pt}
\label{eq:SDC_DUE_compressed}
\end{equation}

\subsection{Results: Global Clipper on Transformer models}
\label{subsec:global_clipper_results}
\begin{figure}[t]
\centering     %%% not \center
\subfigure[Evaluating fault mitigation performance using $IVMOD_{fd}$ metric]{\label{fig:dino_detr_sdc}\includegraphics[width=\linewidth]{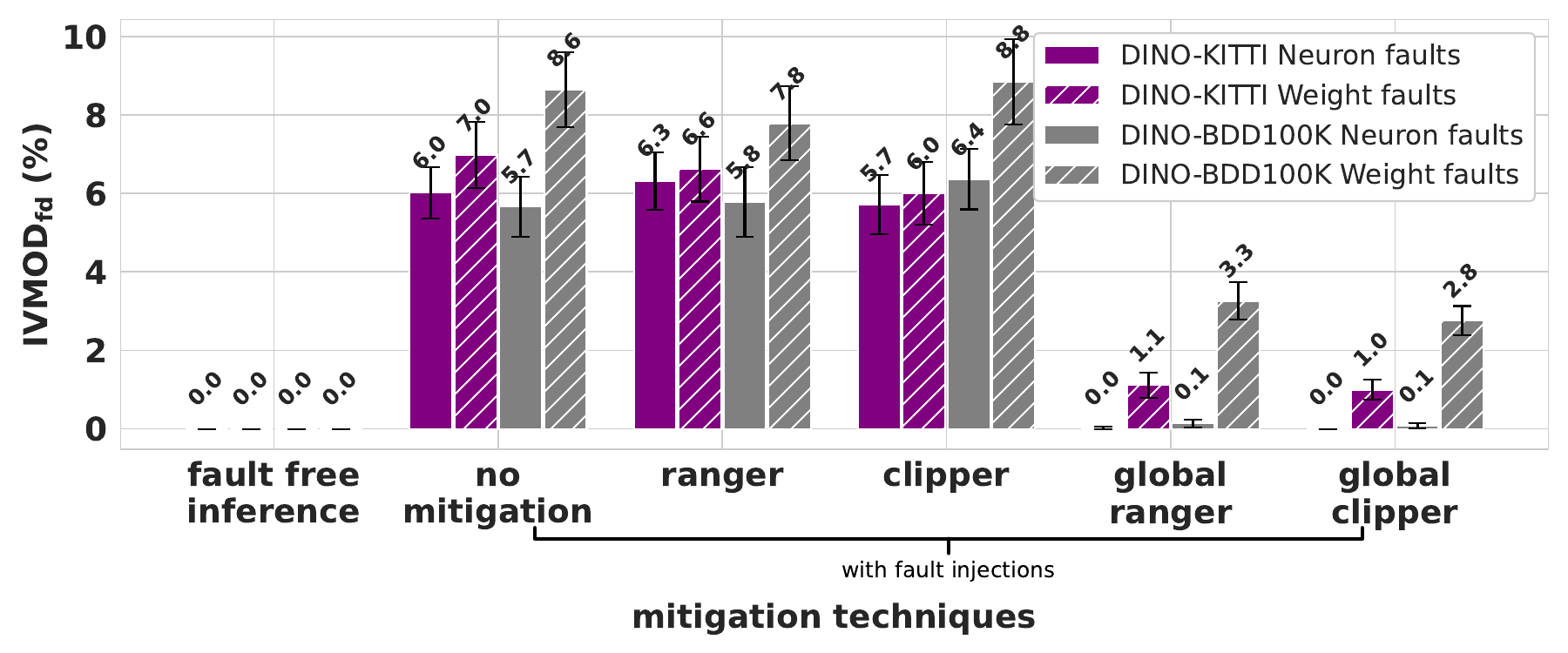}}
\subfigure[Evaluating fault mitigation performance using Average Precision metric]{\label{fig:dino_detr_ap50}\includegraphics[width=\linewidth]{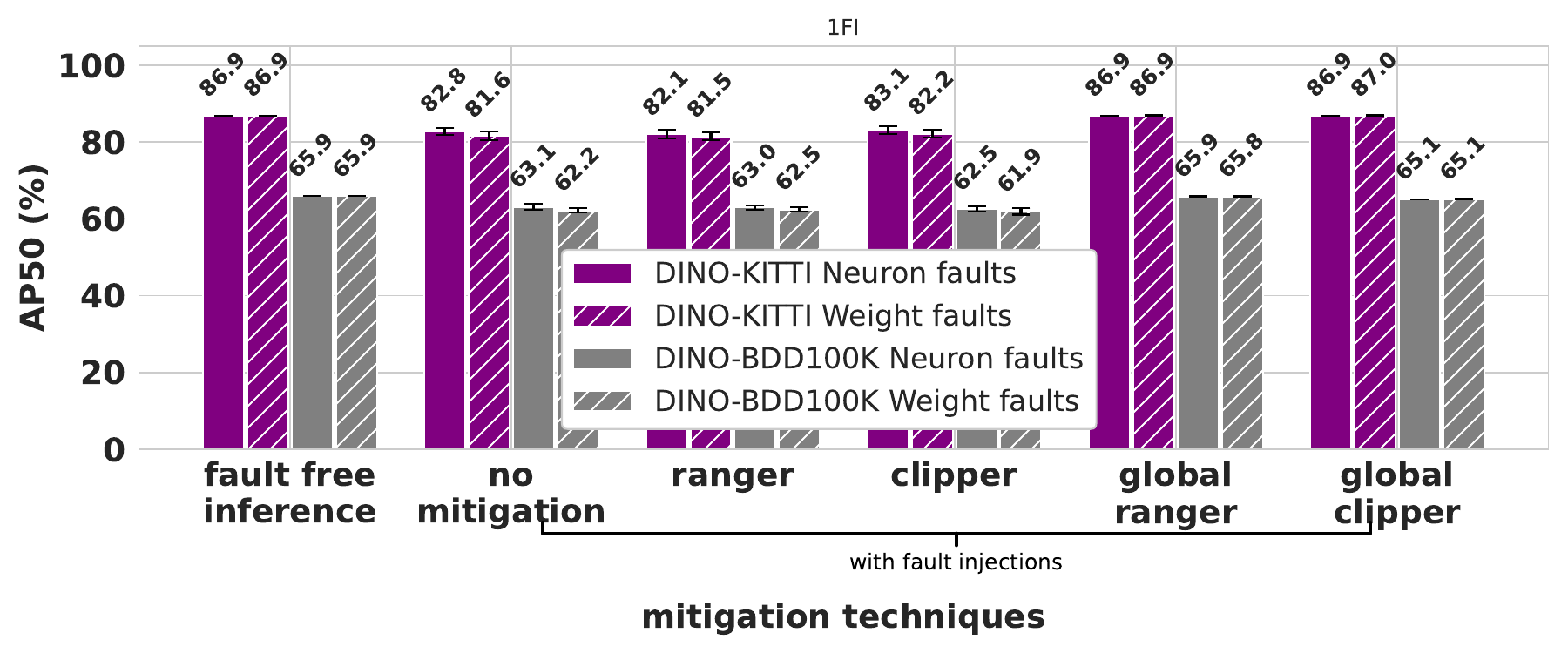}}
\caption{Evaluation of Global Clipper on DINO-DETR Transformer models.}
% \vspace{-0.45cm} 
\label{fig:dino_detr_sdc_ap50}
\end{figure}

\begin{figure}[t]
\centering     %%% not \center
\subfigure[Evaluating fault mitigation performance using $IVMOD_{fd}$ metric]{\label{fig:lite_detr_sdc}\includegraphics[width=\linewidth]{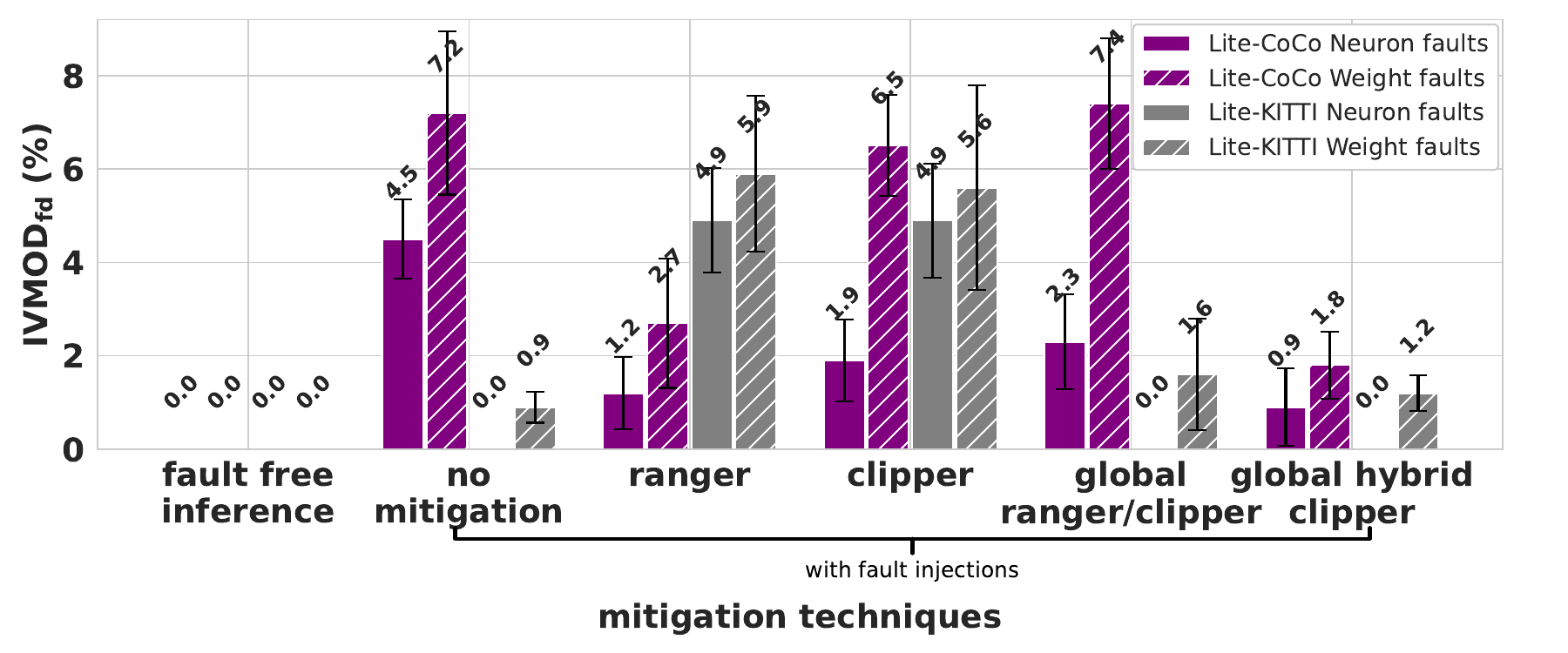}}
% \vspace{-0.22cm} 
\subfigure[Evaluating fault mitigation performance using Average Precision metric]{\label{fig:lite_detr_ap50}\includegraphics[width=\linewidth]{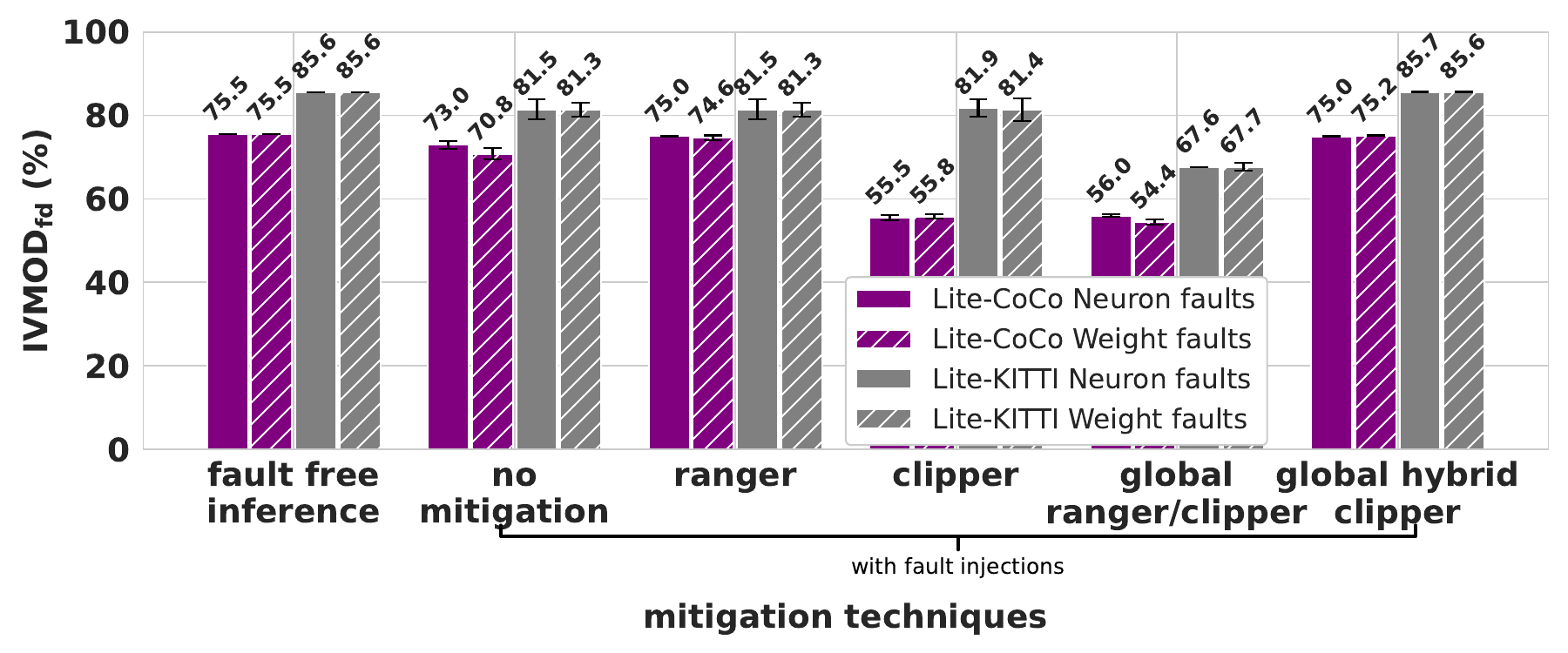}}
\caption{Evaluation of Global Clipper on Lite-DETR models.}
\label{fig:lite_detr_sdc_ap50}
% \vspace{-0.45cm}
\end{figure}

This section showcases the superior mitigation capability of Global Clipper and Global Hybrid Clipper over existing solutions like Ranger and Clipper. This is demonstrated on DINO-DETR and Lite-DETR models with various datasets injected with single bit-flip errors (see \cref{fig:dino_detr_sdc_ap50} and \cref{fig:lite_detr_sdc_ap50}), we also investigate vulnerability across datasets, including layer-wise fault injection experiments.

The \cref{fig:dino_detr_sdc_ap50} illustrates the model's vulnerability and evaluates the effectiveness of the various range restriction techniques in mitigating faults using $IVMOD_{fd}$ and  AP50 metrics. This includes results from experiments involving fault injections into weights and neurons. The Global Clipper significantly outperforms the existing solutions like Ranger and Clipper in mitigating the impact of the faults. 
The Global Clipper performs better in mitigating faults occurring in neurons, reducing faulty detections to nearly 0\%, and for weight faults, the vulnerabilities are reduced to less than $\sim$ 3\% by outperforming other state-of-the-art algorithms like Ranger and Clipper. 
Additionally, as shown in \cref{fig:dino_detr_ap50}, a single bit flip in the inferences notably impacts the AP50 metric. For instance, the AP50 of DINO-DETR trained on KITTI decreases from 86.9 to 81.6 when injected with weight faults, and the Global Clipper effectively restores the AP50 to its original accuracy. 
In the context of DINO-DETR, Global Ranger and Global Clipper demonstrate similar mitigation performance. However, their performance is not as expected when applied to Lite-DETR due to the unique transformer architecture based on DINO-DETR. This presents an interesting scenario, leading to the introduction of the Global Hybrid Clipper. The Global Hybrid Clipper merges Global Clipper and Ranger layers. In this setup, Global Clipper is applied to Activation layers, while Global Ranger is used for the linear layers within self-attention blocks. In this scenario, the hybrid version of Global Clipper restores performance accuracy to baseline and reduces faulty detections significantly, as shown in \cref{fig:lite_detr_sdc_ap50}. For instance, it decreases from 4.5\% to 0.5\% in the case of neuron faults.

\begin{figure}[b]
  \centering
  \includegraphics[width=\linewidth]{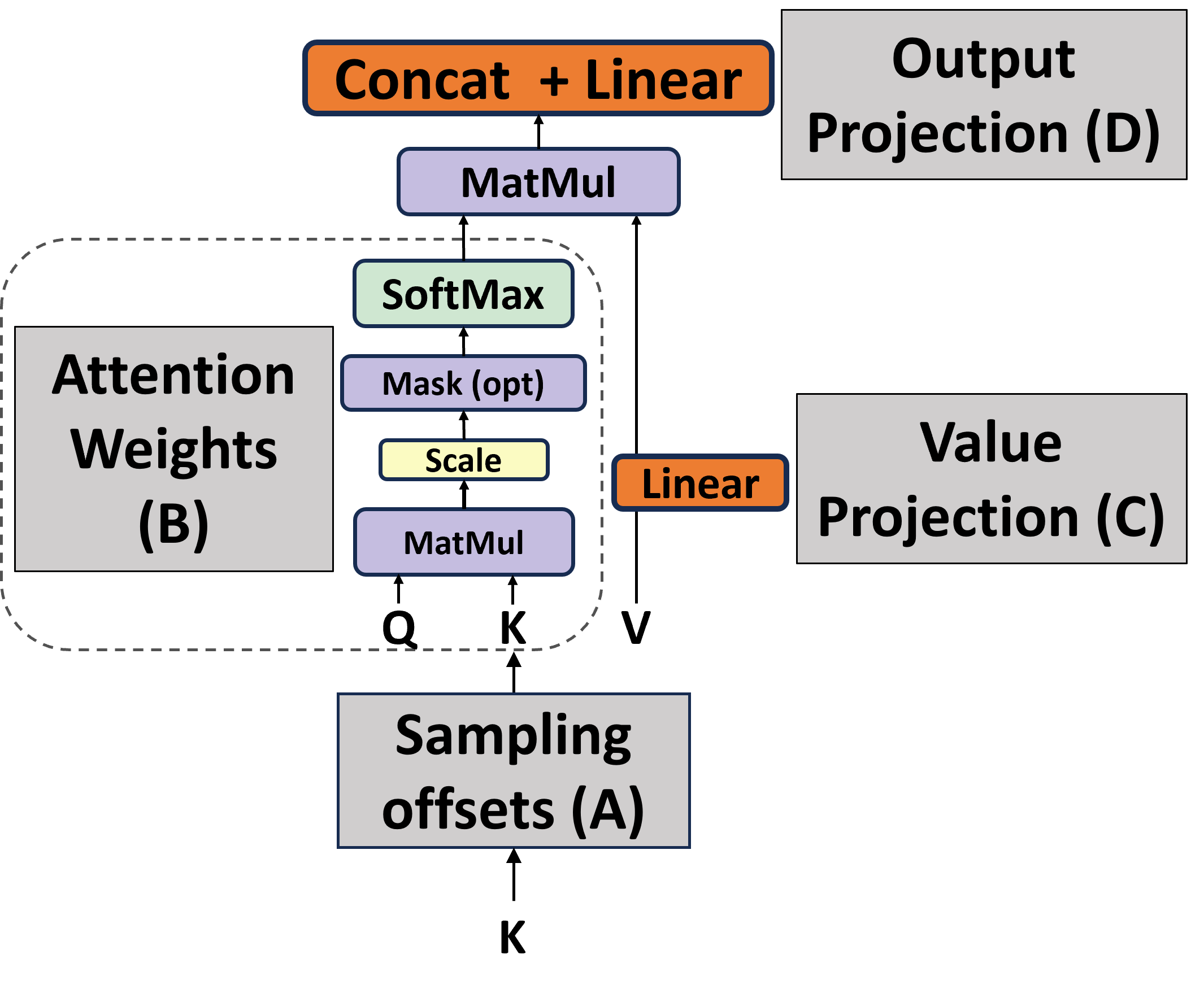}
  \caption{Individual components of Encoders and Decoders (Attention Blocks)}
  \vspace{-0.45cm}
  \label{fig:attention_block}
\end{figure}

\begin{figure*}[t]
\centering     %%% not \center
\subfigure[]{\label{fig:transform_1FI}\includegraphics[width=\linewidth]{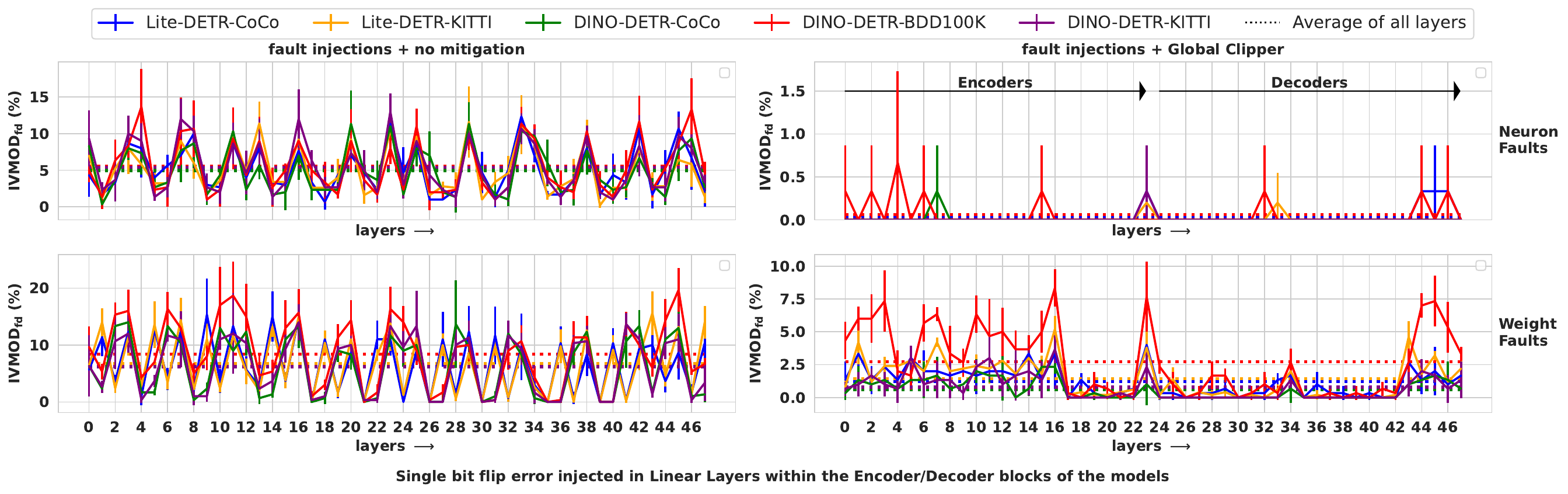}}
\vspace{-0.35cm} 
\subfigure[]{\label{fig:CNN_1FI}\includegraphics[width=\linewidth]{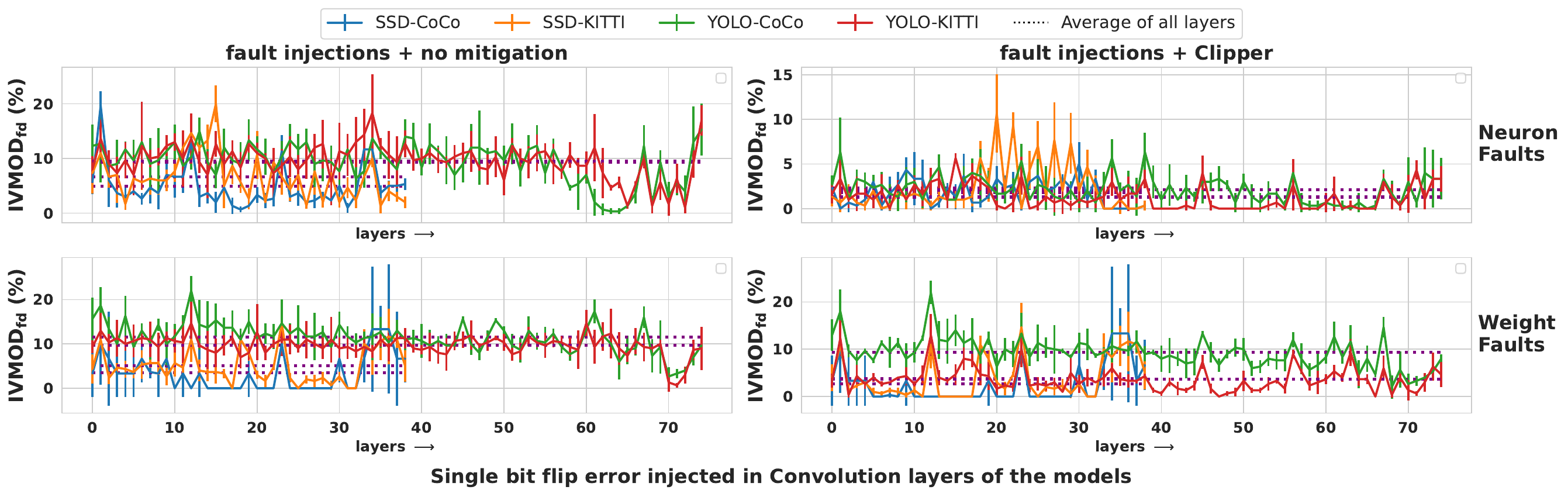}}
% \vspace{-0.38cm} 
\caption{Layer-wise impact of Single-bit flip error on vulnerability metric based on faulty detections on transformers and CNN-based object detection models.}
\label{fig:single_bit_flip_sdc}
\end{figure*}

\begin{figure*}[b]
\centering     %%% not \center
\subfigure[]{\label{fig:transform_10FI}\includegraphics[width=\linewidth]{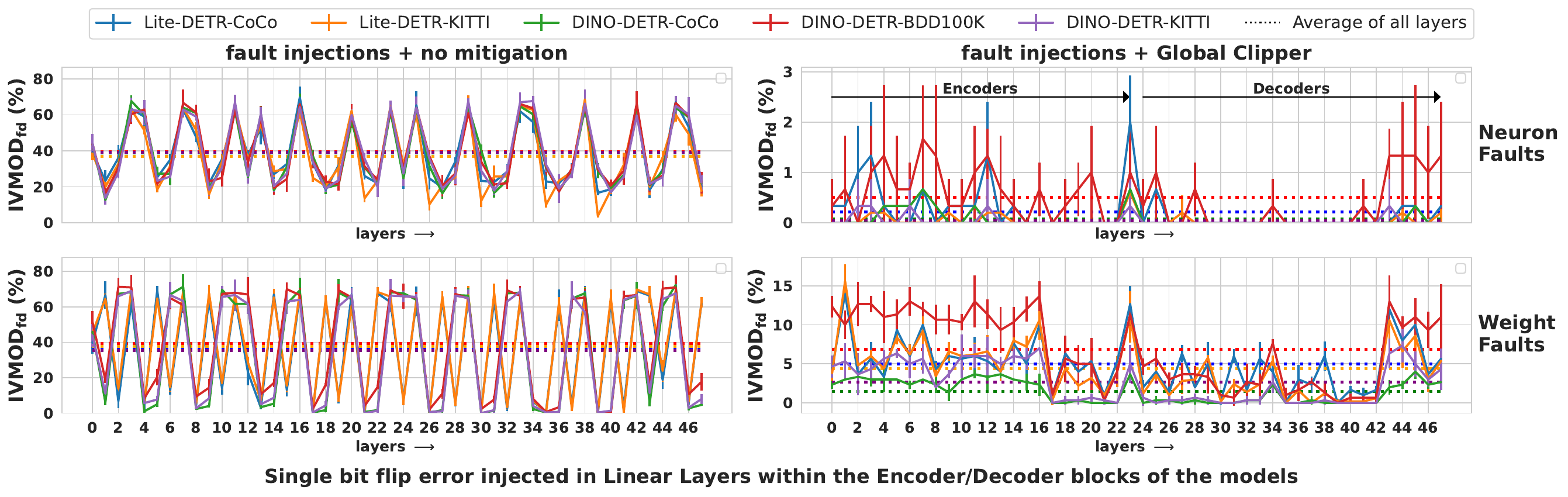}}
\vspace{-0.35cm} 
\subfigure[]{\label{fig:CNN_10FI}\includegraphics[width=\linewidth]{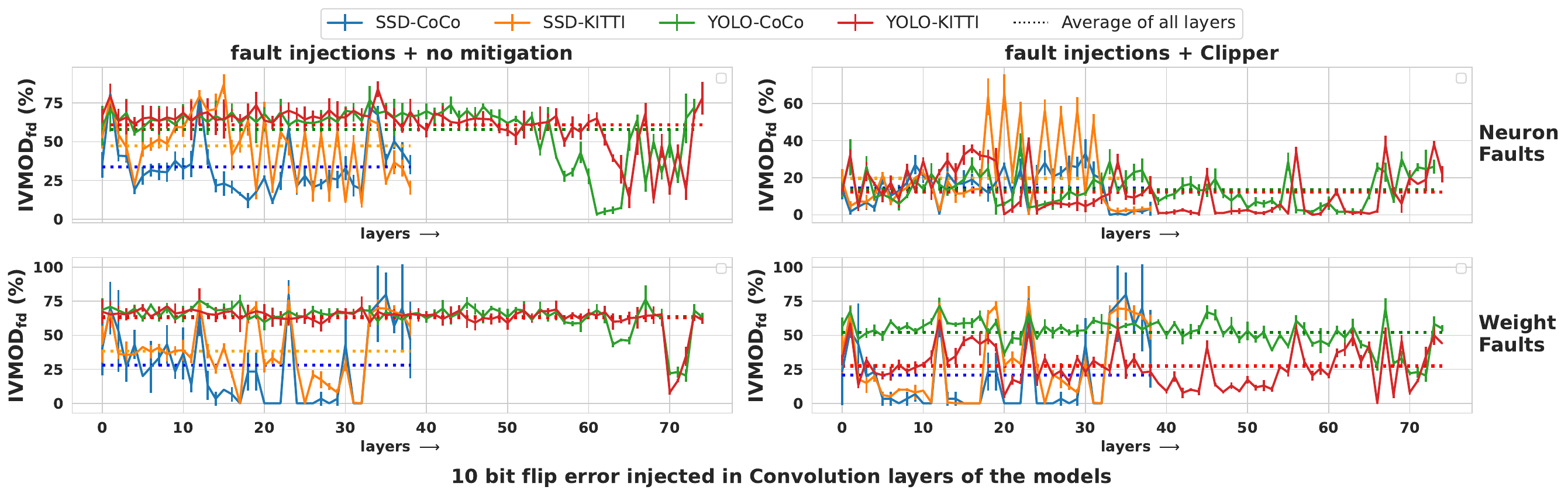}}
% \vspace{-0.38cm} 
\caption{Layer-wise impact of 10-bit flip error on vulnerability metric based on faulty detections on transformers and CNN-based object detection models.}
\label{fig:multi_bit_flip_sdc}
\end{figure*}

% \input{sec_new/ablation}

% \subsection{Ablation Study: Vulnerability and Resiliency Analysis of Attention and Convolution Layers}
\subsection{Ablation Study}
\label{subsec:ablation_study}
\subsubsection{Vulnerability and Resiliency Analysis of Attention and Convolution Layers}

This section outlines the ablation study, including single-bit and multiple-bit (10-bit) flip experiments on CNN and transformer-based object detection models. Results are shown in \cref{fig:single_bit_flip_sdc} and \cref{fig:multi_bit_flip_sdc}. Global Clipper and Global Hybrid Clipper were employed for transformers, while Clipper was used for CNNs due to its superior performance over Ranger (see \cref{fig:CNN_1FI} and \cref{fig:CNN_10FI}).

\begin{figure}[ht]
\centering     %%% not \center
\subfigure[]{\label{fig:attention_block_vulnerability_n}\includegraphics[width=\linewidth]{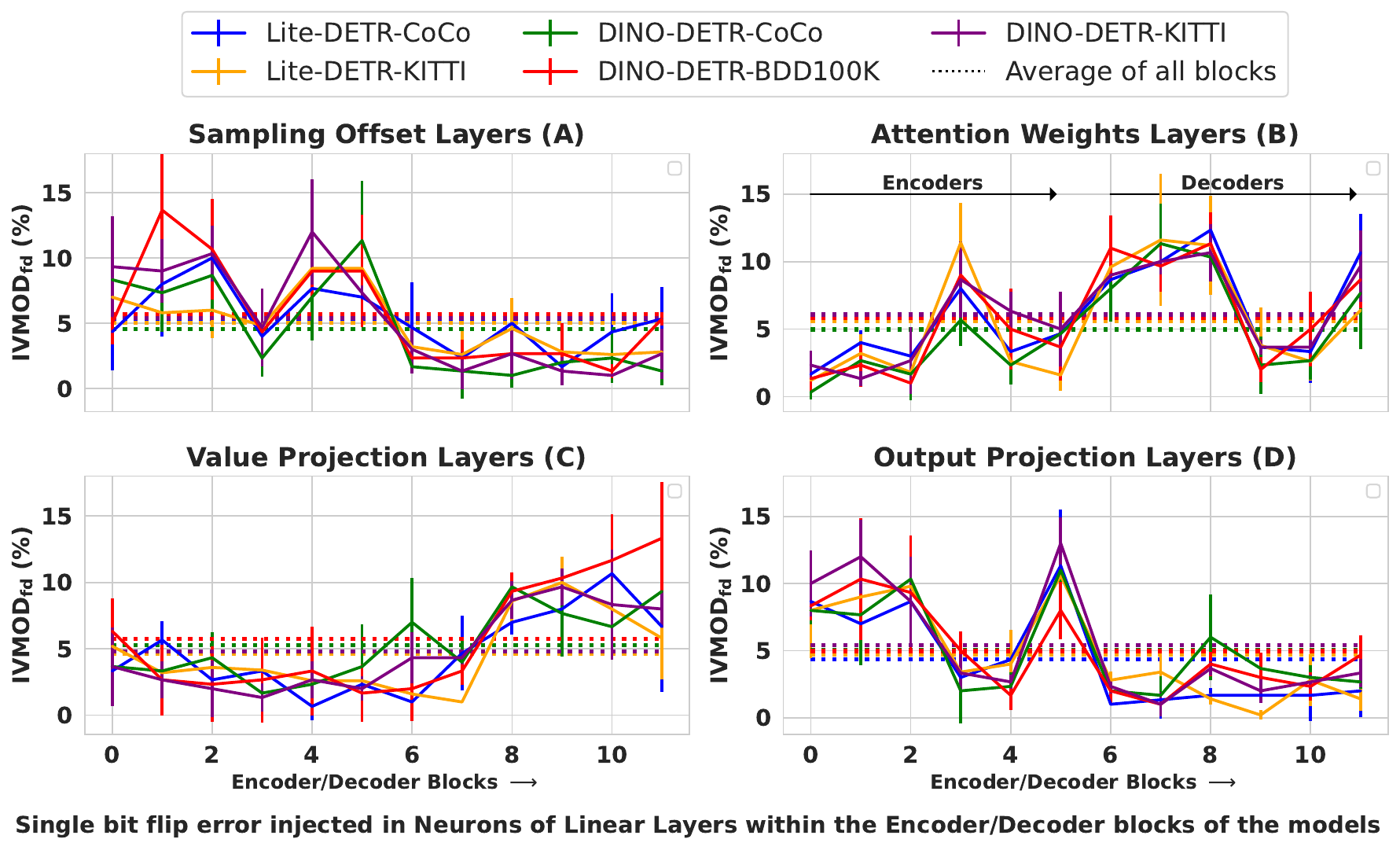}}
\subfigure[]{\label{fig:attention_block_vulnerability_w}\includegraphics[width=\linewidth]{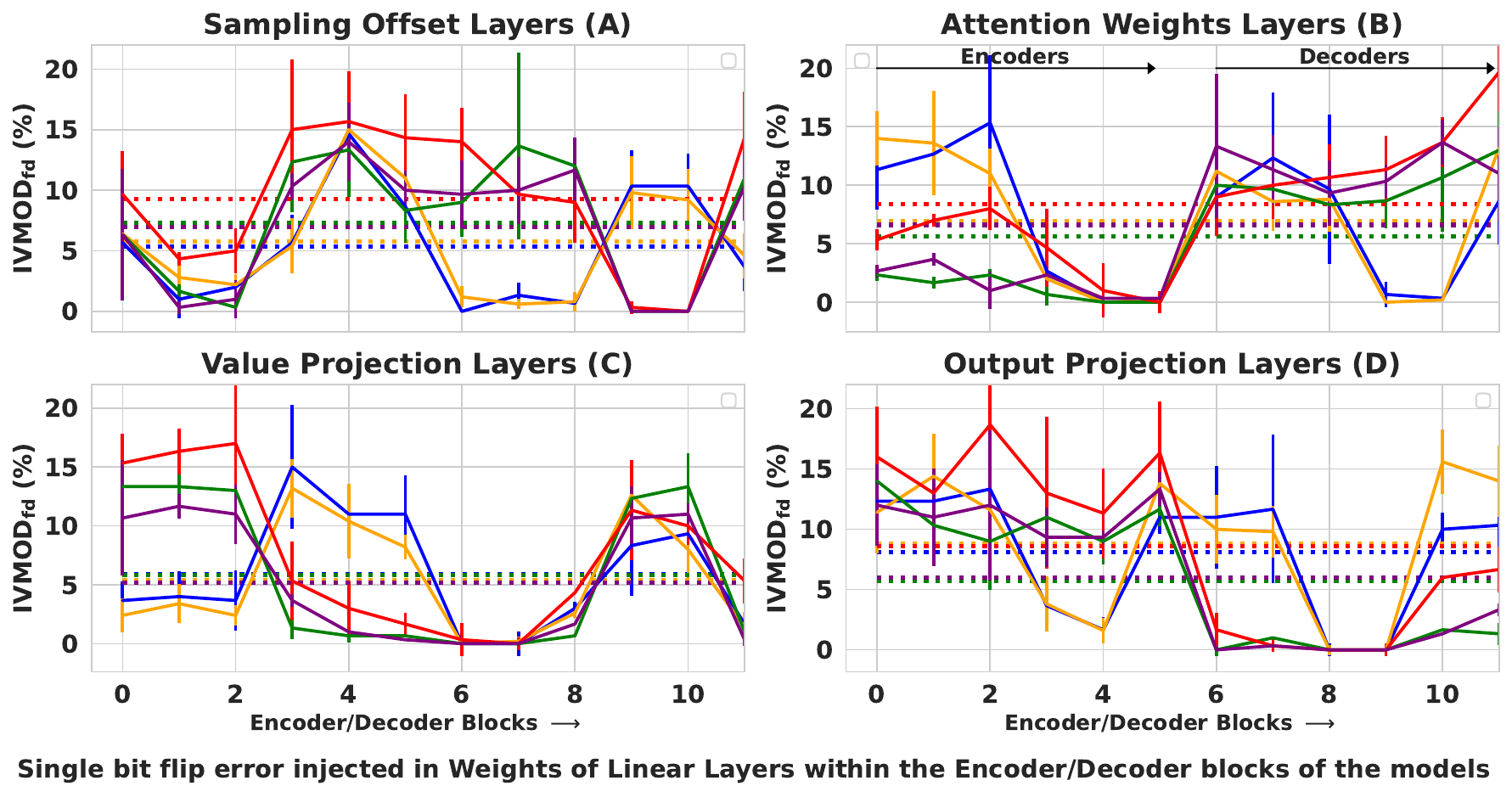}}
% \vspace{-0.38cm}
\caption{Impact of Single-bit flip error on individual linear layers of the Attention Blocks.}
\label{fig:attention_block_vulnerability}
\vspace{-0.45cm} 
\end{figure}

Each plot represents 10,000 inferences, highlighting Global Clipper's superior handling of bit-flip errors. Comparing CNN and self-attention layers reveals that transformers generally exhibit greater fault-injection resilience than CNNs. For instance, DINO-DETR and Lite-DETR consist of six encoders and decoders, each with four linear layers (as depicted in \cref{fig:attention_block}), resulting in 48 layers per model. These transformer layers exhibit distinct characteristics in response to fault injections compared to CNNs. These observations are consistent across both single-bit and multi-bit flip experiments (\cref{fig:single_bit_flip_sdc} and \cref{fig:multi_bit_flip_sdc}). Depending on the transformer model variant, encoders and decoders are interchangeable with self-attention blocks.

CNNs show no discernible pattern across layers, suggesting uniform susceptibility to generating faulty detections under bit-flip errors. In contrast, transformer models display greater inherent resilience. With suitable mitigation techniques, transformers offer enhanced safety against soft errors compared to CNN-based object detection models, making them preferable for applications demanding robustness against bit-flip errors.

\begin{figure}[t]
\centering     %%% not \center
\subfigure[Neuron faults]{\label{fig:attention_block_vulnerability_DF_KDA_neurons}\includegraphics[width=\linewidth]{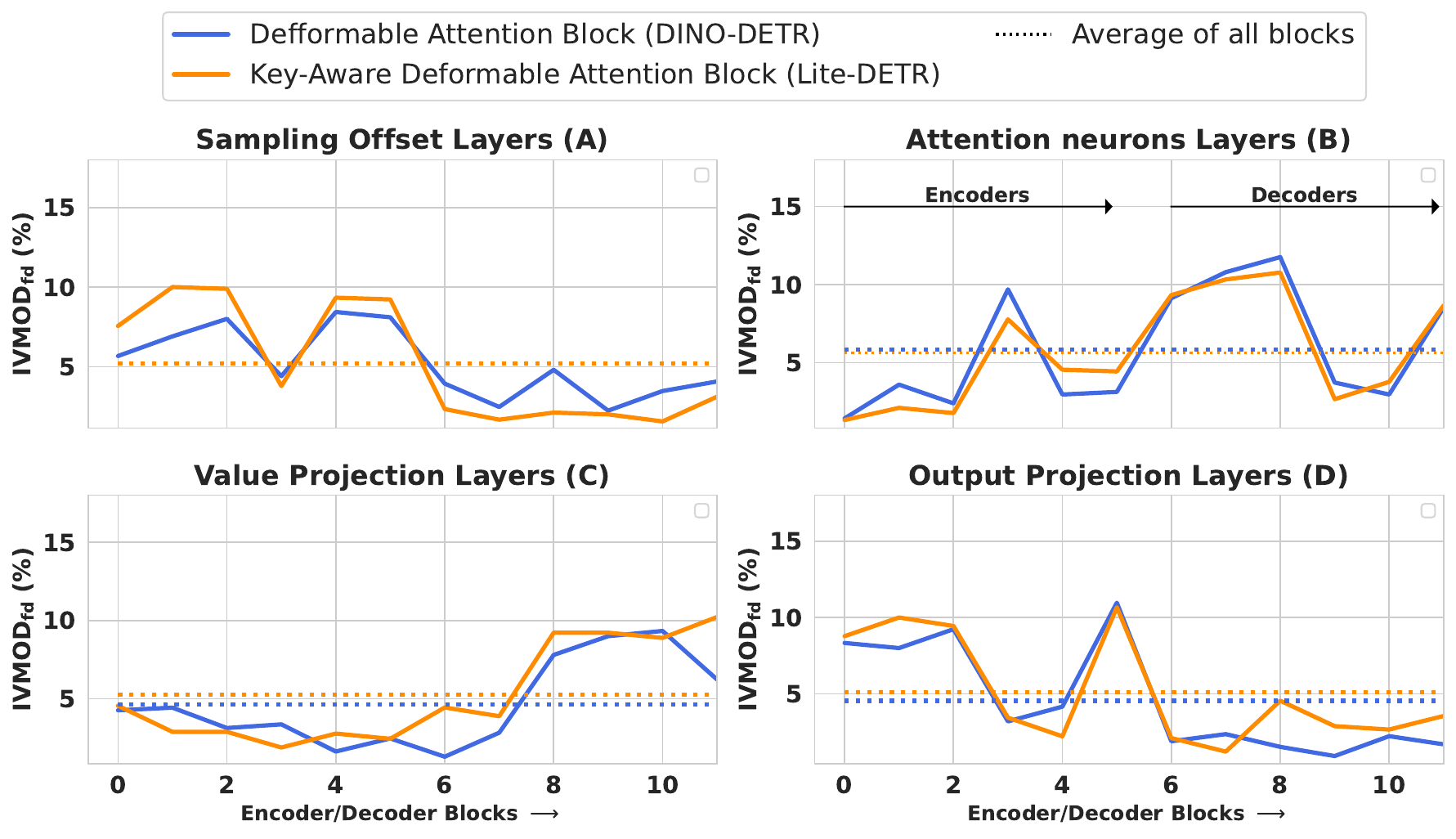}}

\subfigure[Weight faults]{\label{fig:attention_block_vulnerability_DF_KDA_weights}\includegraphics[width=\linewidth]{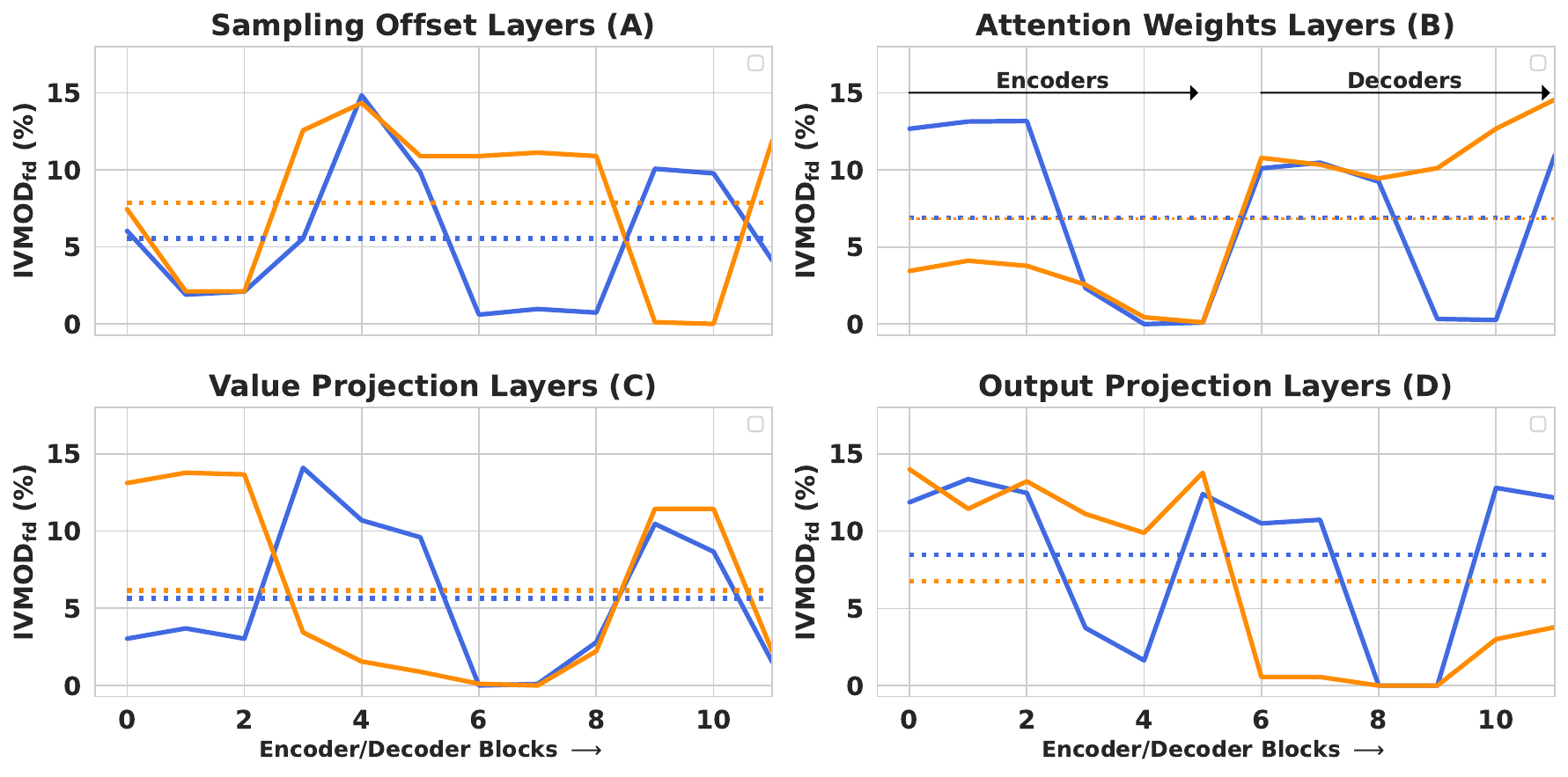}}
% \vspace{-0.38cm}
\caption{Comparison of vulnerability in Deformable Attention (DF) of DINO-DETR and Key-Aware Deformable (KDA) Attention in Lite-DETR under neuron and weight faults. The vulnerability metric ($IVMOD_{fd}$), shown in \cref{fig:attention_block_vulnerability}, averaged across datasets, emphasizing differences between DF and KDA.}
\label{fig:attention_block_vulnerability_DF_KDA}
\vspace{-0.45cm} 
\end{figure}

\begin{figure}[b]
  \centering
  \includegraphics[width=\linewidth]{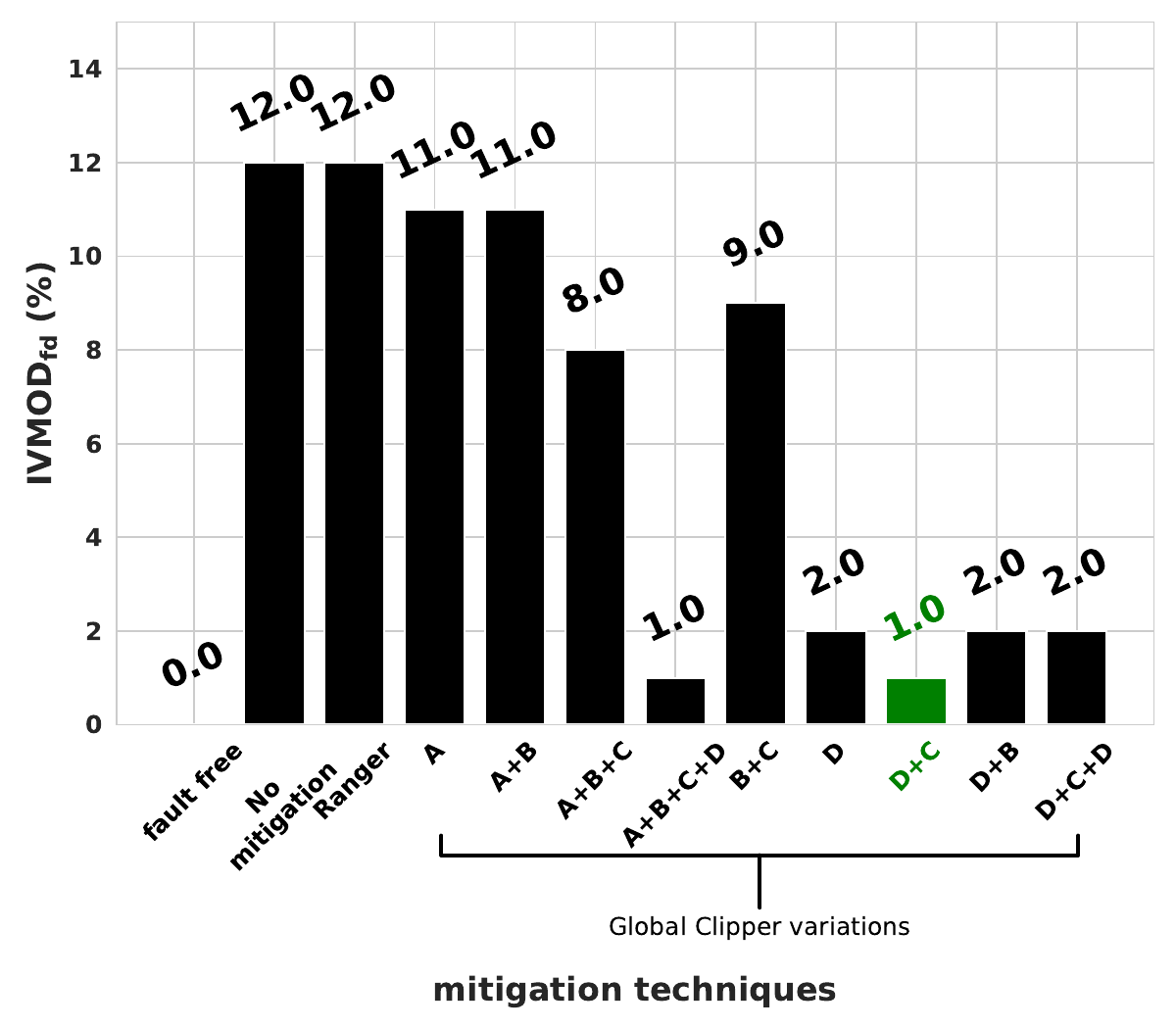}
  \caption{Integrating Global Clipper in Attention Block with minimal overhead.}
  \vspace{-0.45cm} 
  \label{fig:global_clipper_placement}
\end{figure}

To better understand the transformer's vulnerability, we analyzed single-bit flip errors in attention block linear layers A, B, C, and D (see \cref{fig:attention_block}). Data from \cref{fig:transform_1FI} and \cref{fig:transform_10FI} was segmented into four plots in \cref{fig:attention_block_vulnerability}, illustrating each layer’s vulnerability across encoders and decoders in 12 Attention Blocks. Layers such as Sampling Offset, Attention Weights, Value Projection, and Output Projection showed consistent vulnerability across CoCo, KITTI, BDD100K datasets, and DINO-DETR and Lite-DETR models for neuron faults. Variations in vulnerability due to weight faults depend on the attention block type, as shown in \cref{fig:attention_block_vulnerability_DF_KDA}. DINO-DETR uses Deformable Attention (DF) Layers, while Lite-DETR employs Key-Aware Deformable Attention (KDA) Layers, enhancing efficiency and attention mapping. The effects of weight faults on DF and KDA blocks vary, influencing encoders and decoders differently (\cref{fig:attention_block_vulnerability_DF_KDA_weights}), but neuron fault vulnerability remains stable across datasets and models (\cref{fig:attention_block_vulnerability_DF_KDA_neurons}). These observations underscore two key transformer traits in vulnerability, distinct from CNNs:

\begin{itemize}
  \item The vulnerability estimation of a transformer model's layers demonstrates consistent characteristics across different datasets during inference, unlike CNNs.
  \item The model's vulnerability estimation is influenced by the Self-Attention Block variant used in the model and remains consistent across different architectures and datasets during inference.
\end{itemize}
Thorough vulnerability analysis can greatly improve online safety and risk management, assisting in dynamic risk assessment and model monitoring across the lifecycle of deployed transformer-based models. Our findings demonstrate that despite continual learning and weight adjustments, these model's vulnerabilities remain stable, ensuring robust and reliable AI systems are maintained.
% \textcolor{red}{Add a brief example of ways to use these vulnerability characteristics of transformers during online safety assessment or ways to feed this information to the scheduler to execute more vulnerable layers in protective circuits and run the remaining less vulnerable layers with minimal protection}

\subsubsection{Integrating Global Clipper with minimal additional overhead}
As depicted in \cref{fig:intro_global_clipper}, the Global Clipper layers are integrated into four linear layers within the Attention Block (see \cref{fig:attention_block}). The efficiency of adding these layers can be further evaluated by conducting a simple experiment that examines the impact of each Global Clipper Layer. This experiment entails the injection of a single-bit flip error at the initial stage of the Attention Block, particularly at the Sampling Offsets layer (stage A). Following this, the Global Clipper layer is selectively activated at different combinations of stages A, B, C, and D within the four linear layers of the Attention Block, as detailed in \cref{fig:attention_block}.
Our observations indicate that integrating Global Clipper solely into the Output Projection and Value Projection layers (D and C in Figure \ref{fig:attention_block}), in addition to other activation layers, yields mitigation performance comparable to that achieved by incorporating it into all linear layers, as illustrated in Figure \ref{fig:global_clipper_placement}
\section{Conclusion}
\label{sec:conclusion}
This study introduces Global Clipper and Global Hybrid Clipper to enhance the safety of transformer-based object detection models in critical settings, effectively minimizing faulty inferences to nearly ~ 0\%. We evaluated these solutions by conducting fault injection campaigns with transformer and CNN models across three datasets, totalling approximately 3.3 million inferences. Our extensive experiments and findings indicate that transformer models exhibit better inherent resilience to soft errors than CNN models. Evaluating these solutions provides insights into their effectiveness in real-world applications, contributing significantly to model robustness and computer vision safety. Future research should explore these solutions in transformer-based semantic segmentation and video tracking models to further enhance safety. 

% \section{Appendices}

% If your work needs an appendix, add it before the
% ``\verb|\end{document}|'' command at the conclusion of your source
% document.

% Start the appendix with the ``\verb|\appendix|'' command:
% \begin{lstlisting}
% \appendix
% \end{lstlisting}
% and note that in the appendix, sections are lettered, not
% numbered. 

% \begin{acknowledgments}
%   Thanks to the developers of ACM consolidated LaTeX styles
%   \url{https://github.com/borisveytsman/acmart} and to the developers
%   of Elsevier updated \LaTeX{} templates
%   \url{https://www.ctan.org/tex-archive/macros/latex/contrib/els-cas-templates}.  
% \end{acknowledgments}

\begin{acknowledgments}
This work was partially funded by the Federal Ministry for Economic Affairs and Energy of Germany as part of the research project SafeWahr (Grant Number: 19A21026C).
\end{acknowledgments}
%%
%% Define the bibliography file to be used
\bibliography{sample-ceur}

\appendix


% The sources for the ceur-art style are available via
% \begin{itemize}
% \item \href{https://github.com/yamadharma/ceurart}{GitHub},
% % \item \href{https://www.overleaf.com/project/5e76702c4acae70001d3bc87}{Overleaf},
% \item
%   \href{https://www.overleaf.com/latex/templates/template-for-submissions-to-ceur-workshop-proceedings-ceur-ws-dot-org/pkfscdkgkhcq}{Overleaf
%     template}.
% \end{itemize}

\end{document}